%% file: neurips_2026.tex
\documentclass{article}

    \PassOptionsToPackage{numbers, compress}{natbib}

 \usepackage[main, final]{neurips_2026}

\usepackage[utf8]{inputenc} 
\usepackage[T1]{fontenc}    
\usepackage{hyperref}       
\usepackage{url}            
\usepackage{booktabs}       
\usepackage{amsfonts}       
\usepackage{nicefrac}       
\usepackage{microtype}      
\usepackage{subcaption}
\usepackage{xcolor, tabularx}
\definecolor{cmarkgreen}{HTML}{2E7D32}
\definecolor{xmarkred}{HTML}{C62828}
\definecolor{oursblue}{HTML}{E3F2FD}
\newcommand{\cmark}{{\color{cmarkgreen}\ding{51}}}
\newcommand{\xmark}{{\color{xmarkred}\ding{55}}}
\usepackage{graphicx}
\usepackage{multirow}
\usepackage{comment}
\usepackage{wrapfig}
\usepackage{amsmath}
\usepackage{colortbl} 
\usepackage{svg}
\usepackage[capitalize]{cleveref}
\usepackage{pifont}

\usepackage{tikz}
\usepackage{pgfplots}
\usepgfplotslibrary{groupplots}
\pgfplotsset{compat=newest}

\title{No Pose, No Problem in 4D: Feed-Forward Dynamic Gaussians from  Unposed Multi-View Videos}

\author{%
Matteo Balice\textsuperscript{1}\hspace{0.7em}Yanik Künzi\textsuperscript{2}\hspace{0.7em}Chenyangguang Zhang\textsuperscript{2}\hspace{0.7em}Matteo Matteucci\textsuperscript{1}\\
\textbf{Marc Pollefeys}\textsuperscript{2}\hspace{0.7em}\textbf{Sungwhan Hong}\textsuperscript{2,3}\thanks{Corresponding author}\\[5pt]
\textsuperscript{1}Politecnico di Milano,\ \textsuperscript{2}ETH Z\"urich \textsuperscript{3}ETH AI Center\\[5pt]
\tt\ \small \textcolor{blue!60!black}{\url{https://bralani.github.io/nopo4d_html/}}
\vspace{-10pt}
}

\definecolor{backbone_color}{HTML}{4A4A7A}
\definecolor{frozen_color}{HTML}{6BA38A}
\definecolor{trainable_color}{HTML}{E07A5F}

\begin{document}

\renewcommand{\thefootnote}{\fnsymbol{footnote}}%
\setcounter{footnote}{1}
\maketitle

\renewcommand{\thefootnote}{\arabic{footnote}}%
\input{chapters/0_abstract}
\input{chapters/1_introduction}
\input{chapters/2_related-work}
\input{chapters/3_method}
\input{chapters/4_experiments}
\input{chapters/5_discussion}
\input{chapters/6_acknowledgement}
\input{chapters/7_future-work}
\input{chapters/8_bibliography}

\appendix
\input{chapters/appendix}


\newpage

\input{checklist.tex}
\end{document}

%% file: chapters/0_abstract.tex
\begin{abstract}
Recent feed-forward 3D gaussian splatting methods have made dramatic progress on individual aspects of 3D scene reconstruction, but no existing method jointly addresses dynamic content, multi-view input, and unknown camera poses in a single feed-forward pass. Methods that handle dynamics either require accurate camera poses or accept only monocular input; pose-free multi-view methods address only static scenes; and per-scene optimization methods bridge some of these gaps but at minutes-to-hours cost per scene. We introduce NoPo4D, the first feed-forward system that addresses this empty quadrant. Building on a pretrained geometry backbone and recent 4D Gaussian frameworks, NoPo4D introduces a velocity decomposition that splits Gaussian motion into per-pixel image-plane shifts and depth changes, allowing direct supervision from pseudo ground-truth optical flow on the 2D component. This sidesteps both the differentiable rendering that couples prior posed methods to pose accuracy and the 3D motion ground truth that prior pose-free methods require. The system is rounded out by a bidirectional motion encoder for cross-view and cross-frame feature aggregation, and view-dependent opacity that mitigates cross-view and cross-timestep Gaussian misalignments. On four multi-view dynamic benchmarks, NoPo4D consistently outperforms prior feed-forward baselines, and with an optional post-optimization stage surpasses per-scene optimization methods, while running orders of magnitude faster. 
\end{abstract}

%% file: chapters/1_introduction.tex
\section{Introduction}
\label{sec:intro}

Reconstructing dynamic 3D scenes from a handful of time-synchronized video streams, without known camera poses, without per-scene optimization, and in a single forward pass, would bring free-viewpoint rendering of everyday activities to practical scale. Applications range from sports and performance capture with minimal infrastructure, to egocentric scene understanding~\cite{chen2026prose,zhang2026chronograph,sun2026lime}, AR~\cite{ccelen2026genrec}, robotics~\cite{han2026geometric}, to immersive content creation outside controlled studio environments. Yet achieving this in practice demands solving three problems simultaneously: modeling dynamic content, fusing information across viewpoints, and estimating geometry without known camera poses.
 
Recent feed-forward methods have achieved remarkable progress along individual axes of this problem. Static reconstruction from unposed multi-view images can now be performed in under a second~\cite{hong2024unifying,lin_depth_2025, wang_vggt_2025,veicht2026zipsplat}, while feed-forward 4D methods have begun to tackle dynamic scenes from monocular video~\cite{xu_4dgt_2025, yang_neoverse_2026}. However, each existing approach relaxes only a subset of the target constraints, whether by requiring known camera poses, accepting only monocular input, restricting to specific domains such as driving, or relying on per-scene optimization that takes minutes to hours~\cite{wang_monofusion_2025, wang_shape_2024}. As shown in Tab.~\ref{tab:capability}, no existing feed-forward method jointly handles dynamic scenes from unposed, multi-view video of general environments.
 
We present \textbf{NoPo4D}, the first feed-forward system that addresses this empty quadrant. Filling it requires solving three problems that none of the constituent settings face in isolation. First, motion supervision must operate without GT poses or 3D motion annotations, neither of which is available at scale. Prior feed-forward dynamic methods sidestep one constraint at the cost of the other: posed approaches supervise through differentiable rendering~\cite{xu_4dgt_2025}, while pose-free approaches supervise velocities directly against 3D motion ground truth~\cite{yang_neoverse_2026}. We avoid both by decomposing Gaussian motion into per-pixel image-plane shifts and depth changes, supervising the image-plane component directly with pseudo ground-truth optical flow. Second, motion must remain consistent across both cameras and consecutive frames, which we address with a bidirectional motion encoder that performs joint self-attention across views and timesteps. Third, geometric inconsistencies in pose-free reconstruction now compound across cameras \emph{and} timesteps, a more severe problem than either monocular dynamic or static multi-view methods face; we mitigate this with view-dependent opacity.
 
We evaluate NoPo4D on four multi-view dynamic benchmarks: ExoRecon~\cite{wang_monofusion_2025}, Immersive Light Field~\cite{broxton_immersive_2020}, Kubric~\cite{greff_kubric_2022}, and N3DV~\cite{li2022neural}. Across all four, NoPo4D consistently outperforms prior feed-forward baselines. With optional post-optimization, NoPo4D surpasses per-scene optimization methods. Ablations validate that each design choice contributes meaningfully.
In summary, our contributions are as follows:
\begin{itemize}
    \item We identify the setting of feed-forward, pose-free, multi-view dynamic scene reconstruction and show via systematic comparison that no prior method addresses it.
    \item We introduce NoPo4D, the first feed-forward system that operates in this setting. Our key technical contribution is a velocity decomposition that splits Gaussian motion into per-pixel image-plane shifts and depth changes, allowing optical flow to supervise the image-plane component directly. We complement this with a bidirectional motion encoder that aggregates features across views and frames, and adopt view-dependent opacity to handle cross-view and cross-timestep Gaussian misalignments.
    \item We evaluate on four multi-view dynamic benchmarks, demonstrating that NoPo4D consistently outperforms feed-forward baselines and, with optional post-optimization, surpasses per-scene optimization methods while running orders of magnitude faster.
\end{itemize}
 

\begin{table}[t]
\centering
\caption{\textbf{Capability comparison of related methods.} NoPo4D is the first feed-forward method that jointly handles dynamic scenes from unposed,  multi-view video of general environments. \textbf{FF}: feed-forward inference. \textbf{Unposed}: no ground-truth camera poses required. \textbf{MV}: supports multi-view input. \textbf{Dyn.}: models dynamic scenes. \textbf{General}: not restricted to a specific domain.}
\label{tab:capability}
\footnotesize
\renewcommand{\arraystretch}{1.15}
\begin{tabular*}{\textwidth}{@{\extracolsep{\fill}}lcccccc@{}}
\toprule
\textbf{Method} & \textbf{Type} & \textbf{FF} & \textbf{Unposed} & \textbf{MV} & \textbf{Dyn.} & \textbf{General} \\
\midrule
NoPoSplat~\cite{ye_no_2024} & Static & \cmark & \cmark & \cmark & \xmark & \cmark \\
FLARE~\cite{zhang_flare_2025} & Static & \cmark & \cmark & \cmark & \xmark & \cmark \\
AnySplat~\cite{jiang_anysplat_2025} & Static & \cmark & \cmark & \cmark & \xmark & \cmark \\
\addlinespace[2pt]
4DGT~\cite{xu_4dgt_2025} & Dyn. & \cmark & \xmark & \xmark & \cmark & \cmark \\
NeoVerse~\cite{yang_neoverse_2026} & Dyn. & \cmark & \cmark & \xmark & \cmark & \cmark \\
DGGT~\cite{dggt} & Dyn. & \cmark & \cmark & \cmark & \cmark & \xmark \\
\addlinespace[2pt]
Shape of Motion~\cite{wang_shape_2024} & Opt. & \xmark & \xmark & \xmark & \cmark & \cmark \\
MonoFusion~\cite{wang_monofusion_2025} & Opt. & \xmark & \xmark & \cmark & \cmark & \cmark \\
\midrule
\textbf{NoPo4D (Ours)} & \textbf{Dyn.} & \cmark & \cmark & \cmark & \cmark & \cmark \\
\bottomrule
\end{tabular*}
\end{table}\vspace{-10pt}
 

%% file: chapters/2_related-work.tex
\section{Related Work}

\paragraph{Static Feed-Forward Reconstruction.}
The DUSt3R family~\cite{wang_dust3r_2023, leroy_grounding_2024} pioneered feed-forward 3D reconstruction by regressing dense point maps from image pairs, bypassing classical multi-stage pipelines. This idea was extended to dynamic point maps~\cite{zhang_monst3r_2024, han_d2ust3r_2025}, incremental multi-view processing~\cite{wang20253d, wang2025continuous}, and parallel many-view inference~\cite{yang_fast3r_2025}. VGGT~\cite{wang_vggt_2025} scaled the paradigm to hundreds of views with joint prediction of cameras, depth, and point tracks, while Depth Anything 3 (DA3)~\cite{lin_depth_2025} showed that a plain DINOv2~\cite{oquab_dinov2_2023} backbone with alternating attention achieves spatially consistent multi-view geometry with fewer parameters. A parallel thread predicts Gaussian primitives directly from images. PixelSplat~\cite{charatan_pixelsplat_2023} and MVSplat~\cite{chen_mvsplat_2024} demonstrated feed-forward Gaussian prediction but require known poses. Removing this assumption, NoPoSplat~\cite{ye_no_2024} fine-tunes a MASt3R-style architecture for pose-free prediction, C3G~\cite{an2025c3g} builds on top of the existing foundation models~\cite{wang_vggt_2025}, and subsequent methods generalize to cascaded pose-geometry pipelines~\cite{zhang_flare_2025}, unconstrained views~\cite{jiang_anysplat_2025, smart_splatt3r_2024}, variable-length sequences~\cite{chen2024pref3r, hong_pf3plat_2024}, and unified pose-free and pose-dependent settings~\cite{ye2025yonosplat}. All these methods are restricted to static scenes.\vspace{-5pt}

\paragraph{Feed-Forward Dynamic Reconstruction.}
L4GM~\cite{ren_l4gm_2024} proposed the first feed-forward 4D model but is limited to object-centric scenes with per-frame Gaussians. 4DGT~\cite{xu_4dgt_2025} introduced temporally coherent 4D Gaussians with lifespan and velocity attributes, trained on monocular posed video. NeoVerse~\cite{yang_neoverse_2026} and MoVieS~\cite{lin2026movies} extend this to unposed and motion-aware monocular settings respectively, but both remain restricted to single-camera input. DGGT~\cite{dggt} operates in the unposed multi-view regime but is tailored to driving scenarios. UFO-4D~\cite{hur2026ufo} achieves unposed feed-forward 4D reconstruction from a stereo pair. NoPo4D is, to the best of our knowledge, the first method to jointly handle dynamic scenes, multi-view input, unknown camera poses, and general scene content in a single feed-forward pass.\vspace{-5pt}

\paragraph{Optimization-Based Dynamic Reconstruction.}
Per-scene optimization has been approached via canonical deformation fields~\cite{pumarola_d-nerf_2020, park_nerfies_2021, tretschk_non-rigid_2021} and explicit spatiotemporal factorizations~\cite{cao_hexplane_2023, fridovich-keil_k-planes_2023}. Within Gaussian splatting, one family attaches per-Gaussian motion models: deformation MLPs~\cite{yang_deformable_2023, wu_4d_2024}, per-timestep parameters with rigidity constraints~\cite{luiten_dynamic_2023}, sparse control points~\cite{huang2024sc}, motion scaffolds~\cite{lei_mosca_2024}, and learned trajectories~\cite{stearns2024dynamic}. Another lifts Gaussians to native 4D primitives by conditioning position and opacity on time~\cite{yang_real-time_2024, duan_4d-rotor_2024}. For sparse-view input, Shape-of-Motion~\cite{wang_shape_2024} proposes SE(3) motion bases, an idea also adopted by MonoFusion~\cite{wang_monofusion_2025}, which reconstructs each view independently with monocular depth priors~\cite{yang_depth_2024-1, yang_depth_2024} before aligning them into a shared 4D representation. All these methods require known camera poses and minutes-to-hours of computation per scene.

%% file: chapters/3_method.tex
\section{Method}
\label{sec:method}

\subsection{Problem Formulation}
\label{sec:problem}

Consider $C$ uncalibrated, time-synchronized video streams of a scene, given as images $\mathcal{I} = \left\{(\mathbf{I}^c_t)^T_{t=1}\right\}^C_{c=1}$, where $\mathbf{I}^c_t \in \mathbb{R}^{H \times W \times 3}$ is the frame from camera $c$ at timestep $t$, and each stream contains $T$ frames. We assume the cameras form a static rig, as in capture setups such as Ego-Exo4D~\cite{grauman_ego-exo4d_2024}. NoPo4D jointly predicts: 1) per-camera intrinsics $\hat{\mathbf{K}}^c$ and extrinsics $(\hat{\mathbf{R}}^c, \hat{\mathbf{t}}^c)$, obtained by averaging the backbone's per-frame predictions across $t$ for each camera $c$, and 2) a collection of $G$ 4D Gaussians
\begin{equation}
\left(\boldsymbol{\mu}_g, \boldsymbol{\Sigma}_g, \alpha_g, \mathbf{c}_g, \tau_g, l_g, \mathbf{v}^+_g, \mathbf{v}^-_g, \boldsymbol{\omega}^+_g, \boldsymbol{\omega}^-_g\right)^G_{g=1},
\end{equation}
where each Gaussian carries static attributes mean $\boldsymbol{\mu}_g$, covariance $\boldsymbol{\Sigma}_g$, opacity $\alpha_g$, and color $\mathbf{c}_g$. Following~\cite{xu_4dgt_2025,yang_neoverse_2026}, dynamic attributes include temporal center $\tau_g$, lifespan $l_g$, and forward/backward linear and angular velocities $\mathbf{v}^\pm_g, \boldsymbol{\omega}^\pm_g$ that permit asymmetric motion around the temporal center.

\subsection{Model Architecture}

\begin{figure}[t]
    \centering
    \includegraphics[width=\linewidth]{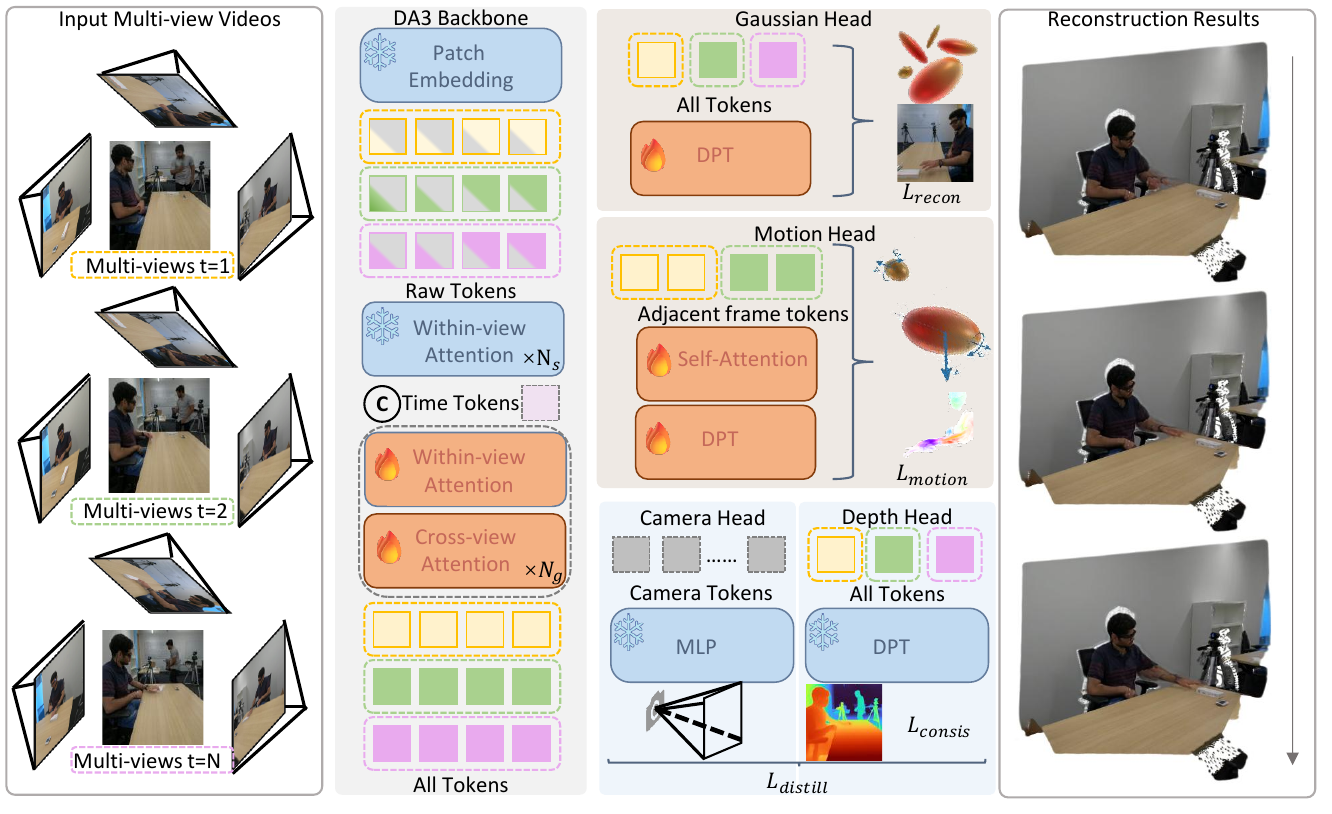}
    \caption{\textbf{Architecture overview.} Given $C$ streams of time-synchronized video, DA3~\cite{lin_depth_2025} first extracts multi-view features through alternating within-view and cross-view attention layers. Pretrained, frozen depth and camera heads then recover per-frame geometry, which is unprojected into Gaussian means $\boldsymbol{\mu}$. Subsequently, two trainable heads decode the remaining attributes: a Gaussian head predicts static parameters ($\mathbf{R}, \mathbf{s}, \alpha, \mathbf{c}$), while a bidirectional motion branch processes consecutive-frame features to produce forward and backward linear and angular velocities ($\mathbf{v}^\pm, \boldsymbol{\omega}^\pm$), a per-pixel motion gate $\rho$, and the temporal covariance $\sigma_g$. All outputs are composed into a single 4D Gaussian representation and rendered via differentiable rasterization.}
    \label{fig:architecture}\vspace{-10pt}
\end{figure}

Fig.~\ref{fig:architecture} illustrates the architecture: a pretrained geometry backbone extracts multi-view features that are routed to frozen camera and depth heads for per-frame geometry and to trainable Gaussian and motion heads for the remaining static and dynamic attributes.

\paragraph{Feature Backbone.}
We employ the pretrained transformer backbone $\mathcal{T}$ from Depth Anything 3~\cite{lin_depth_2025}. Given input frames $\mathcal{I}$, the backbone processes them through a stack of frozen within-view self-attention layers. We inject sinusoidal temporal tokens encoding each frame's timestamp before the trainable alternating within-view and cross-view attention layers, providing the backbone with an explicit temporal signal. This yields multi-view features $\mathcal{F} = \left\{(\mathbf{F}^c_t)^T_{t=1}\right\}^C_{c=1} = \mathcal{T}(\mathcal{I})$, where $\mathbf{F}^c_t \in \mathbb{R}^{N \times D}$ has $N$ tokens per frame and embedding dimension $D$. Frozen pretrained heads then predict per-frame depth maps $\hat{\mathbf{D}}^c_t \in \mathbb{R}^{H \times W}_+$, intrinsics $\hat{\mathbf{K}}^c_t$, and extrinsics $(\hat{\mathbf{R}}^c_t, \hat{\mathbf{t}}^c_t)$; under our static-rig assumption (Sec. \ref{sec:problem}) these are averaged across $t$ to yield the per-camera estimates. Keeping these heads frozen prevents overfitting and empirically yields more stable training (see supplementary material). \vspace{-5pt}

\paragraph{Gaussian Means.}
Each depth map $\hat{\mathbf{D}}^c_t$ is unprojected into a world-space point map $\hat{\mathbf{P}}^c_t \in \mathbb{R}^{H \times W \times 3}$ via the inverse projection $\Pi^{-1}$ that lifts each pixel into a 3D point using its predicted depth and the camera parameters:
\begin{equation}
\hat{\mathbf{P}}^c_t = \Pi^{-1}(\hat{\mathbf{D}}^c_t, \hat{\mathbf{R}}^c, \hat{\mathbf{t}}^c, \hat{\mathbf{K}}^c).
\end{equation}
We instantiate one Gaussian per pixel per frame: each Gaussian is uniquely indexed by $g = (c, t, x, y)$, and its mean $\boldsymbol{\mu}_g$ is set to the corresponding point-map entry $\hat{\mathbf{P}}^c_t(x, y)$. \vspace{-5pt}

\paragraph{Static Attributes.}
A DPT decoder~\cite{ranftl_vision_2021} applied to the backbone features $\mathbf{F}^c_t$ predicts the remaining static parameters per pixel. The covariance $\boldsymbol{\Sigma}_g$ is parameterized by a unit quaternion $\mathbf{q}_g \in \mathbb{R}^4$ encoding rotation and a scale vector $\mathbf{s}_g$, following the standard 3DGS factorization~\cite{kerbl_3d_2023}. Color $\mathbf{c}_g$ is decoded by the same head. We parameterize opacity with spherical harmonic coefficients $\alpha_g \in \mathbb{R}^{(k+1)^2}$ of order $k$ rather than the scalar opacity used in standard 3DGS, making opacity view-dependent. This is critical in our pose-free multi-view dynamic setting: the SH coefficients act as a learned confidence metric that compensates for cross-view and cross-timestep Gaussian misalignments arising from imperfect geometry prediction, allowing unreliable Gaussians to become transparent from problematic viewpoints. \vspace{-5pt}

\paragraph{Decomposed Motion Parameterization.}
We decompose Gaussian motion into per-pixel image-plane shifts and a depth change, which allows direct supervision from 2D optical flow without requiring 3D motion ground truth or a pose-dependent rendering step. This contrasts with prior feed-forward dynamic methods~\cite{xu_4dgt_2025, yang_neoverse_2026}, which predict 3D Gaussian velocities directly. A DPT head decodes per-pixel 2D shifts $(\Delta x, \Delta y)$ and a depth displacement $\Delta d$, with pixel shifts bounded via $\tanh$ and scaled to image dimensions $(W, H)$. For pixel $(x, y)$ in frame $t$ with depth $\hat{\mathbf{D}}^c_t(x, y)$, these define the projected location and depth at $t+1$:
\begin{equation}
\label{eq:pixel_shift}
x' = x + \Delta x, \quad y' = y + \Delta y, \quad \hat{\mathbf{D}}^c_{t+1}(x,y) = \hat{\mathbf{D}}^c_t(x,y) + \Delta d.
\end{equation}
Using the predicted intrinsics $\hat{\mathbf{K}}^c$ with focal lengths $(f_x, f_y)$ and principal point $(c_x, c_y)$, we unproject the source and displaced pixels into camera space and take their difference:
\begin{equation}
\label{eq:cam_disp}
\Delta \mathbf{X}_{\text{cam}} = \begin{pmatrix} \dfrac{(x' - c_x)\, \hat{\mathbf{D}}^c_{t+1} - (x - c_x)\, \hat{\mathbf{D}}^c_t}{f_x} \\[6pt] \dfrac{(y' - c_y)\, \hat{\mathbf{D}}^c_{t+1} - (y - c_y)\, \hat{\mathbf{D}}^c_t}{f_y} \\[6pt] \hat{\mathbf{D}}^c_{t+1} - \hat{\mathbf{D}}^c_t \end{pmatrix}.
\end{equation}
The forward linear velocity $\mathbf{v}^+_g$ for the Gaussian at this pixel is obtained by rotating into world coordinates and scaling by the temporal interval $\Delta t$:
\begin{equation}
\label{eq:velocity}
\mathbf{v}^+_g = \frac{\hat{\mathbf{R}}^c\, \Delta \mathbf{X}_{\text{cam}}}{\Delta t}.
\end{equation}
The backward velocity $\mathbf{v}^-_g$ is computed symmetrically from features of the preceding frame.\vspace{-5pt}

\paragraph{Motion Encoder.}
The displacements above are produced by a bidirectional motion encoder $\mathcal{M}$. For each consecutive pair of frames $(t, t{+}1)$, tokens from all $C$ cameras at frame $t$ are concatenated with tokens from all $C$ cameras at frame $t{+}1$, with two learnable embeddings tagging the source frame versus the neighbour. Joint self-attention over this combined set of $2 C N$ tokens aggregates information across views and time. The resulting tokens are split back into forward and backward halves and decoded by a shared DPT head. Per Gaussian, the head outputs the 2D pixel shifts $(\Delta x, \Delta y)$ and depth displacement $\Delta d$ above, angular velocities $\boldsymbol{\omega}^\pm_g$, a per-pixel motion gate $\rho_g \in (0,1)$ that suppresses motion in static regions, a temporal covariance $\sigma_g$, and a temporal center $\tau_g$ anchored to the source frame's timestamp.\vspace{-5pt}

\paragraph{Post-Optimization.}
We include an optional test-time refinement stage following AnySplat~\cite{jiang_anysplat_2025}. After the feed-forward pass, we prune low-opacity Gaussians, render from input views, and minimize photometric losses between rendered and input images. Since the feed-forward output provides a strong initialization, convergence is rapid.

\subsection{Training}

\paragraph{Losses.}
We optimize using a weighted combination of objectives:
\begin{equation}
\mathcal{L} = \mathcal{L}_{\text{recon}} + \lambda_{\text{motion}}\mathcal{L}_{\text{motion}} + \lambda_{\text{consis}}\mathcal{L}_{\text{consis}} + \mathcal{L}_{\text{distill}}.
\end{equation}

The reconstruction loss $\mathcal{L}_{\text{recon}} = \lambda_{\text{MSE}}\mathcal{L}_{\text{MSE}} + \lambda_{\text{SSIM}}\mathcal{L}_{\text{SSIM}} + \lambda_{\text{LPIPS}}\mathcal{L}_{\text{LPIPS}}$ is evaluated exclusively on rendered target frames. Since the model is pose-free, rendering target views during training requires aligning predicted poses into a common coordinate frame; we use a two-pass strategy with closed-form Sim(3) alignment (details in supplementary).

The motion loss supervises Gaussian dynamics by aligning the predicted pixel shifts $(\Delta x, \Delta y)$ from Eq.~\eqref{eq:pixel_shift} with pseudo ground-truth optical flow $\mathbf{f}^*$ from SEA-RAFT~\cite{wang_sea-raft_2024}. A natural alternative is to render 2D flow from the predicted Gaussian velocities and compare it to $\mathbf{f}^*$, but rendering introduces dependence on Gaussian alignment and amplifies pose-induced errors. Since each Gaussian corresponds one-to-one with an input pixel, the predicted pixel shift directly encodes the source Gaussian's 2D motion, allowing us to bypass rendering entirely. We supervise only pixels with ground-truth shifts exceeding 2 pixels ($\Omega' = \{p \in \Omega \mid \|\mathbf{f}^*(p)\|_2 > 2\}$):
\begin{equation}
\lambda_{\text{motion}}\mathcal{L}_{\text{motion}} = \lambda_{\text{flow}} \frac{1}{|\Omega'|} \sum_{p \in \Omega'} \left\| (\Delta x, \Delta y)(p) - \mathbf{f}^*(p) \right\|_1.
\end{equation}

The depth consistency loss ensures rendered depth matches backbone priors:
\begin{equation}
\lambda_{\text{consis}}\mathcal{L}_{\text{consis}} = \lambda_{\text{consis}} \frac{1}{|\Omega|} \sum_{p \in \Omega} \|\hat{\mathbf{D}}(p) - \mathbf{D}(p)\|^2_2.
\end{equation}

To preserve pretrained knowledge, we distill from the frozen DA3 Giant model using pseudo-labels $\{\mathbf{T}^*, \mathbf{D}^*, \nabla\mathbf{D}^*\}$:
\begin{equation}
\mathcal{L}_{\text{distill}} = \lambda_{\text{pose}}\mathcal{L}_{\text{pose}} + \lambda_{\text{depth}}\mathcal{L}_{\text{depth}} + \lambda_{\text{normal}}\mathcal{L}_{\text{normal}},
\end{equation}
where $\mathcal{L}_{\text{pose}}$ is a Huber loss between predicted and teacher poses, and $\mathcal{L}_{\text{depth}} = \lambda_{\text{depth}}\|\mathbf{D} - \mathbf{D}^*\|^2_2$ and $\mathcal{L}_{\text{normal}} = \lambda_{\text{normal}}\|\nabla\mathbf{D} - \nabla\mathbf{D}^*\|^2_2$.

\paragraph{Training Curriculum.}
We train in two stages. The first stage trains on single-camera viewpoints, optimizing static scene representations and initializing the motion heads. The second stage extends to multi-camera batches, sampling frames distributed across multiple synchronized cameras. Within the second stage, we apply a curriculum on the temporal stride between context frames, linearly increasing it from a small initial value. Starting with closely spaced frames allows the model to reliably learn from optical flow supervision before tackling wider temporal baselines.

%% file: chapters/4_experiments.tex
%

\section{Experiments}
\label{sec:experiments}
\subsection{Implementation Details}
\label{sec:implementation}
 
We train on \textasciitilde{}2{,}900 Ego-Exo4D~\citep{grauman_ego-exo4d_2024} scenes, with frames undistorted and downsampled to $796 \times 448$ (maximum 448 on the longer side); following AnySplat~\citep{jiang_anysplat_2025}, we randomize aspect ratios in $[0.5, 1.0]$ and apply random center-cropping (77--100\%) and horizontal flipping. Training runs in two stages of 20{,}000 steps each: single-camera viewpoints, then 16-frame batches across 1--4 cameras with temporal stride sampled from 4--8 (warmed up over the first 2{,}000 steps). We use Depth Anything 3 Large as backbone. We use AdamW~\citep{loshchilov_decoupled_2017} with learning rates $5 \times 10^{-6}$ (backbone alternating attention), $5 \times 10^{-5}$ (motion encoder), and $1 \times 10^{-4}$ (velocity DPT and Gaussian head), with 1{,}000-step linear warmup and cosine annealing to $\frac{1}{10}$ of the initial rate. Loss weights are $\lambda_{\text{MSE}} = 1$, $\lambda_{\text{LPIPS}} = 0.5$, $\lambda_{\text{SSIM}} = 0.05$, $\lambda_{\text{consis}} = 0.1$, $\lambda_{\text{flow}} = 0.02$, $\lambda_{\text{pose}} = 1.0$, $\lambda_{\text{depth}} = 0.1$, $\lambda_{\text{normal}} = 0.1$. Post-optimization runs 100 steps after pruning Gaussians with opacity below 0.01, taking under 5 seconds per scene. Training uses batch size 1 per GPU at $448 \times 448$ on 4 NVIDIA GH200s.
 
\subsection{Experimental Setup}
\label{sec:setup}
 
\paragraph{Datasets and protocol.}
We evaluate on four multi-view dynamic benchmarks. \textit{ExoRecon}~\citep{wang_monofusion_2025} provides sparse multi-view captures derived from Ego-Exo4D and serves as our in-distribution benchmark; we use the same test scenes as ExoRecon to allow direct comparison with MonoFusion. \textit{Immersive Light Field}~\citep{broxton_immersive_2020} contains real-world dynamic scenes from a multi-camera rig, \textit{Kubric}~\citep{greff_kubric_2022} is a synthetic dataset with rendered scenes of varying complexity, and \textit{N3DV}~\citep{li2022neural} provides multi-view dynamic videos with denser camera coverage that allow evaluation under varying camera counts and from novel viewpoints. The latter three are out-of-distribution relative to our Ego-Exo4D training data, allowing us to assess generalization. We adopt a chunk-based evaluation protocol on all four benchmarks: each scene is partitioned into chunks of 5 consecutive timestamps, where 4 surrounding frames serve as input context and the middle frame is held out as the target. Each method synthesizes the target frame at each input camera viewpoint, and we report PSNR, SSIM, and LPIPS averaged across cameras, chunks, and test scenes. We cap evaluation at 60 chunks per scene for ExoRecon and at 100 chunks per scene for the others. On N3DV, we additionally evaluate from novel viewpoints at moderate ($5.7^\circ$--$27.7^\circ$) and extreme ($34.6^\circ$--$71.9^\circ$) rotations from the input cameras.\vspace{-5pt}

\paragraph{Baselines.}
We compare NoPo4D against feed-forward and per-scene optimization baselines. The feed-forward baselines include \textit{NeoVerse}~\citep{yang_neoverse_2026} and \textit{MoVieS}~\citep{lin2026movies}, both originally designed for monocular video; we extend them to multi-view input by independently reconstructing each camera stream and aligning per-view outputs into a shared coordinate frame. \textit{DGGT}~\citep{dggt} is the closest existing method, operating in the unposed multi-view dynamic regime. The optimization baselines include \textit{MonoFusion}~\citep{wang_monofusion_2025} and \textit{MV-SOM}~\citep{som2024} and \textit{Dyn3D-GS}~\citep{luiten_dynamic_2023}; all require known camera poses, in contrast to NoPo4D. For fair comparison on ExoRecon, we retrain DGGT on the same Ego-Exo4D-derived training data as NoPo4D, eliminating any distribution-shift advantage. The remaining feed-forward baselines are designed for monocular input and cannot be straightforwardly retrained on the multi-view setup; we evaluate them as released. On Immersive Light Field, Kubric, and N3DV, NoPo4D operates outside its training distribution. Note that NeoVerse was trained on Kubric and is therefore in-distribution on that benchmark; all other feed-forward baselines are also out-of-distribution on these test sets.
 
\subsection{Quantitative Evaluation}
\label{sec:quant_evaluation}

\paragraph{In-distribution performance on ExoRecon.}
Table~\ref{tab:exorecon} reports performance on ExoRecon, our in-distribution benchmark. To enable a fair comparison, we also retrain DGGT~\cite{dggt} on the same Ego-Exo4D-derived training data as ours. Overall, we find that NoPo4D substantially outperforms other feed-forward baselines. With an optional 100-step post-optimization, NoPo4D surpasses the per-scene optimization SOTA MonoFusion, and further broadens the gap.  This demonstrates that a strong feed-forward initialization, combined with brief refinement, can match or exceed methods that rely on full per-scene optimization with known calibration.\vspace{-5pt}
 
\begin{table}[htbp]
  \caption{\textbf{Held-out view synthesis on ExoRecon~\citep{wang_monofusion_2025}.} \textbf{FF} denotes feed-forward inference.}
  \label{tab:exorecon}
  \centering
  \footnotesize
  \renewcommand{\arraystretch}{1.1}
  \setlength{\tabcolsep}{6pt}
  \begin{tabular}{@{}lccccc@{}}
    \toprule
    Method & FF & PSNR $\uparrow$ & SSIM $\uparrow$ & LPIPS $\downarrow$ \\
    \midrule
    Dyn3D-GS~\citep{luiten_dynamic_2023}     &    \xmark        & 24.28 & 0.692 & 0.539 \\
    MV-SOM~\citep{som2024}                   &    \xmark        & 26.91 & 0.890 & 0.138 \\
    MonoFusion~\citep{wang_monofusion_2025}  &   \xmark         & {30.43} & \textbf{0.930} & \textbf{0.060} \\\midrule
    \rowcolor{oursblue}
    \textbf{NoPo4D (Ours, post-opt.)}        &   \xmark         & \textbf{31.95} & {0.928} & {0.075} \\
    \midrule
    MoVies~\citep{lin2026movies}            & \cmark & 18.78 & 0.725 & 0.288 \\
    DGGT~\citep{dggt}                        & \cmark & 19.67 & 0.645 & 0.417 \\
    DGGT (EgoExo4D)~\citep{dggt}             & \cmark & 20.44 & 0.704 & 0.418 \\
    NeoVerse~\citep{yang_neoverse_2026}      & \cmark & 20.03 & 0.714 & 0.354 \\
    \midrule
    \rowcolor{oursblue}
    \textbf{NoPo4D (Ours)}                   & \cmark & \textbf{29.15} & \textbf{0.886} & \textbf{0.125} \\
    
    \bottomrule
  \end{tabular}\vspace{-10pt}
\end{table}
 
\paragraph{Generalization across distributions.}
Tables~\ref{tab:ood_generalization} and \ref{tab:n3dv_novel_views} report zero-shot performance on three benchmarks outside NoPo4D's Ego-Exo4D training distribution. All feed-forward baselines, except for Neoverse, also operate out-of-distribution on these test sets, so the gaps reflect how well each model's learned priors transfer; the comparison to DGGT, trained on the same data as NoPo4D, further isolates the architectural contribution. NoPo4D consistently outperforms the strongest feed-forward baseline across all three benchmarks. On N3DV in particular, it leads by nearly +4 PSNR from input cameras and remains competitive at moderate ($5.7^\circ$--$27.7^\circ$) novel viewpoints. At extreme ($34.6^\circ$--$71.9^\circ$) viewpoints, pixel-aligned metrics begin to misrepresent perceptual quality, rewarding methods that hedge with blurry, low-confidence predictions over those that produce sharper but slightly-shifted output. As shown in Fig.~\ref{fig:vis_ext}, NoPo4D produces visually cleaner reconstructions than DGGT under such viewpoints despite a small numerical gap, with NeoVerse and MoVieS showing more severe artifacts. These results indicate that NoPo4D's design transfers across capture modalities (egocentric, real multi-camera, synthetic, and dense multi-view) without any domain-specific tuning. Additional analysis of robustness to varying input view counts is provided in the supplementary material.\vspace{-5pt}
 
\begin{table}[htbp]
  \caption{\textbf{Generalization to out-of-distribution datasets.} NoPo4D was not trained on either benchmark. Note that NeoVerse was trained on Kubric and is therefore in-distribution on that benchmark.}
  \label{tab:ood_generalization}
  \centering
  \footnotesize
  \renewcommand{\arraystretch}{1.1}
  \setlength{\tabcolsep}{6pt}
  \begin{tabular}{@{}lcccccc@{}}
    \toprule
    & \multicolumn{3}{c}{Immersive Light Field~\citep{broxton_immersive_2020}} & \multicolumn{3}{c}{Kubric~\citep{greff_kubric_2022}} \\
    \cmidrule(lr){2-4} \cmidrule(lr){5-7}
    Method & PSNR $\uparrow$ & SSIM $\uparrow$ & LPIPS $\downarrow$ & PSNR $\uparrow$ & SSIM $\uparrow$ & LPIPS $\downarrow$ \\
    \midrule
    MoVies~\citep{lin2026movies}             & 16.12 & 0.650 & 0.410 & 14.49 & 0.720 & 0.260 \\
    NeoVerse~\citep{yang_neoverse_2026}      & 19.83 & 0.680 & 0.360 & 16.86 & 0.500 & 0.520 \\
    DGGT~\citep{dggt}                        & 20.65 & 0.700 & 0.360 & 20.14 & 0.530 & 0.430 \\
    DGGT (EgoExo4D)~\citep{dggt}             & 20.23  & 0.685 & 0.453 & 18.22 & 0.551 & 0.522 \\
    \midrule
    \rowcolor{oursblue}
    \textbf{NoPo4D (Ours)}                   & 21.63 & 0.740 & 0.280 & 23.61 & 0.739 & 0.256 \\
    \rowcolor{oursblue}
    \textbf{NoPo4D (Ours, post-opt.)}        & \textbf{24.04} & \textbf{0.817} & \textbf{0.245} & \textbf{28.17} & \textbf{0.838} & \textbf{0.159} \\
    \bottomrule
  \end{tabular}\vspace{-10pt}
\end{table}
 
\begin{table}[htbp]
  \caption{\textbf{View synthesis on N3DV~\citep{li2022neural}.} We report under three viewpoint configurations: original input cameras, moderate novel views ($5.7^\circ$--$27.7^\circ$ rotation), and extreme novel views ($34.6^\circ$--$71.9^\circ$ rotation).}
  \label{tab:n3dv_novel_views}
  \centering
  \footnotesize
  \renewcommand{\arraystretch}{1.1}
  \setlength{\tabcolsep}{4.5pt}
  \resizebox{\textwidth}{!}{%
  \begin{tabular}{@{}lccccccccc@{}}
    \toprule
    & \multicolumn{3}{c}{Input Cameras} & \multicolumn{3}{c}{Novel Cameras (Moderate)} & \multicolumn{3}{c}{Novel Cameras (Extreme)} \\
    \cmidrule(lr){2-4} \cmidrule(lr){5-7} \cmidrule(lr){8-10}
    Method & PSNR $\uparrow$ & SSIM $\uparrow$ & LPIPS $\downarrow$ & PSNR $\uparrow$ & SSIM $\uparrow$ & LPIPS $\downarrow$ & PSNR $\uparrow$ & SSIM $\uparrow$ & LPIPS $\downarrow$ \\
    \midrule
    MoVies~\citep{lin2026movies}             & 14.10 & 0.556 & 0.505 & 15.49 & 0.457 & 0.502 & 10.42 & 0.156 & 0.492 \\
    NeoVerse~\citep{yang_neoverse_2026}      & 24.50 & 0.789 & 0.313 & 25.34 & 0.833 & 0.274 & 17.11 & 0.659 & 0.221 \\
    DGGT~\citep{dggt}                        & 23.08 & 0.749 & 0.268 & 22.92 & 0.735 & 0.260 & 17.12 & 0.590 & 0.212 \\
    DGGT (EgoExo4D)~\citep{dggt}             & 22.49 & 0.713 & 0.356 & 23.10 & 0.751 & 0.239 & 16.82 & 0.582 & 0.206 \\
    \midrule
    \rowcolor{oursblue}
    \textbf{NoPo4D (Ours)}                   & 28.43 & 0.905 & 0.144 & 26.79 & 0.899 & 0.146 & \textbf{19.36} & \textbf{0.771} & \textbf{0.130} \\
    \rowcolor{oursblue}
    \textbf{NoPo4D (Ours, post-opt.)}        & \textbf{31.79} & \textbf{0.956} & \textbf{0.089} & \textbf{27.66} & \textbf{0.909} & \textbf{0.102} & 19.18 & 0.759 & 0.132 \\
    \bottomrule
  \end{tabular}
  }\vspace{-10pt}
\end{table}
\subsection{Qualitative Evaluation}
\label{sec:qual_evaluation}
 
\begin{figure}[t]
    \centering
    \includegraphics[width=\linewidth]{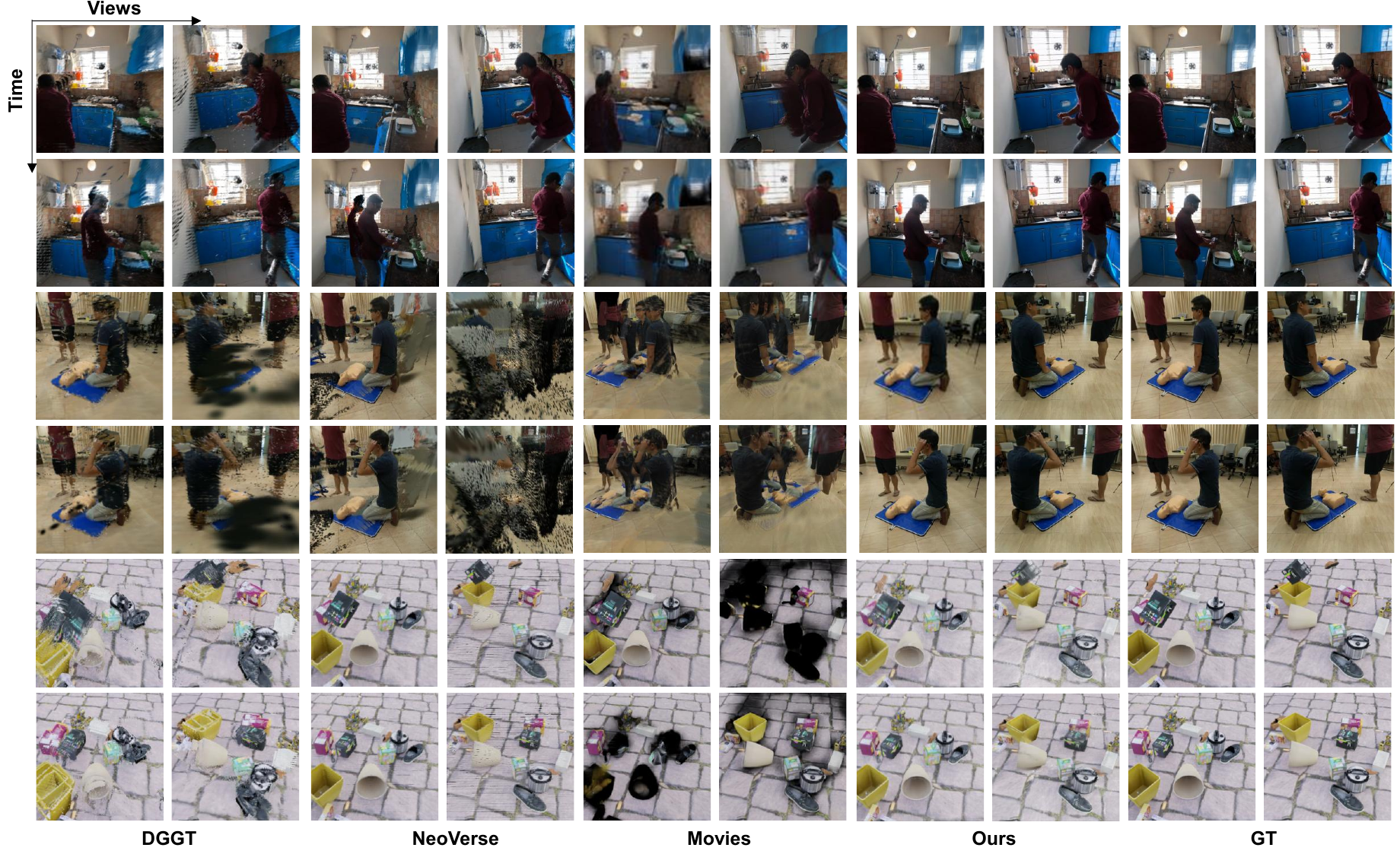}
    \caption{\textbf{Qualitative comparison on ExoRecon~\cite{wang_monofusion_2025} and Kubric~\cite{greff_kubric_2022}.}}
    \label{fig:vis}\vspace{-10pt}
\end{figure}
 
\begin{figure}[t]
    \centering
    \includegraphics[width=\linewidth]{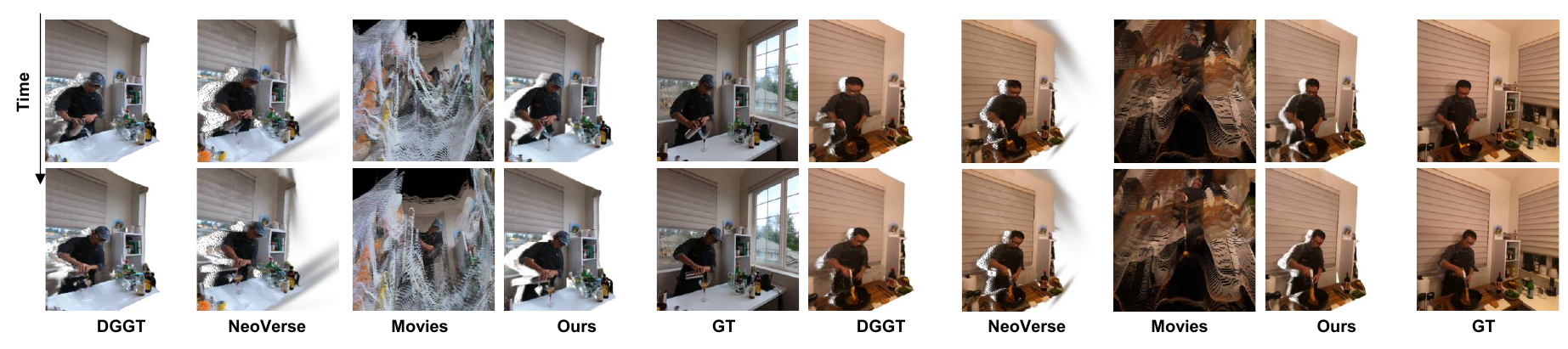}
    \caption{\textbf{Qualitative comparison on N3DV~\cite{li2022neural} under extreme viewpoint changes.}}
    \label{fig:vis_ext}\vspace{-10pt}
\end{figure}
 
Figures~\ref{fig:vis} and~\ref{fig:vis_ext}   compare rendering results across feed-forward methods. NoPo4D recovers fine-scale structure with sharp dynamic foregrounds, while baselines exhibit blurring, geometric distortion, or missing motion. At extreme viewpoints, DGGT shows semi-transparent ghosting and NeoVerse and MoVieS over-smooth, while NoPo4D preserves recognizable geometry and texture, supporting our reading of Table~\ref{tab:n3dv_novel_views}: pixel-aligned metrics under-credit confident-but-shifted predictions relative to blurry-but-aligned ones. We provide video comparisons in the supplementary material. 
\subsection{Ablation Study}
\label{sec:ablation}
 
We conduct ablations to validate each design choice on ExoRecon. Architectural components are isolated in Table~\ref{tab:ablations_architectural}, auxiliary losses in Table~\ref{tab:ablations_loss}. Backbone fine-tuning strategies are analyzed in the supplementary material.
 
\begin{table}[htbp]
  \caption{\textbf{Ablation studies on ExoRecon.} \textit{Left:} architectural component ablation, where each row removes or replaces one component of the full NoPo4D pipeline. \textit{Right:} auxiliary loss ablation, where each column indicates whether a loss is included (\checkmark) or removed (\xmark) from the training objective; the reconstruction loss $\mathcal{L}_{\text{recon}}$ is always active.}
  \label{tab:ablations}
  \centering
  \footnotesize
  \renewcommand{\arraystretch}{1.1}
  \begin{minipage}[t]{0.42\linewidth}
    \centering
    \setlength{\tabcolsep}{4pt}
    \begin{tabular}{@{}lccc@{}}
      \toprule
      Method & PSNR $\uparrow$ & SSIM $\uparrow$ & LPIPS $\downarrow$ \\
      \midrule
      No opacity SH          & 21.61 & 0.758 & 0.247 \\
      No motion branch       & 22.97 & 0.856 & 0.161 \\
      No decomposition       & 27.67 & 0.865 & 0.129 \\
      No temporal encoding   & 27.69 & 0.861 & \textbf{0.125} \\
      \midrule
      \rowcolor{oursblue}
      \textbf{NoPo4D (Ours)} & \textbf{29.15} & \textbf{0.886} & \textbf{0.125} \\
      \bottomrule
    \end{tabular}
    \subcaption{Architectural components.}
    \label{tab:ablations_architectural}
  \end{minipage}%
  \hfill
  \begin{minipage}[t]{0.56\linewidth}
    \centering
    \setlength{\tabcolsep}{3pt}
    \begin{tabular}{@{}cccccccc@{}}
      \toprule
      Recon. & Distill. & Consis. & Flow & PSNR $\uparrow$ & SSIM $\uparrow$ & LPIPS $\downarrow$ \\
      \midrule
      \checkmark & \xmark      & \xmark      & \xmark      & 25.09 & 0.829 & 0.162 \\
      \checkmark & \xmark      & \checkmark  & \xmark      & 3.51  & 0.238 & 0.803 \\
      \checkmark & \xmark      & \checkmark  & \checkmark  & 25.91 & 0.815 & 0.159 \\
      \checkmark & \checkmark  & \xmark      & \checkmark  & 28.02 & 0.867 & \textbf{0.122} \\
      \checkmark & \checkmark  & \checkmark  & \xmark      & 28.19 & 0.869 & 0.125 \\
      \midrule
      \rowcolor{oursblue}
      \checkmark & \checkmark  & \checkmark  & \checkmark  & \textbf{29.15} & \textbf{0.886} & 0.125 \\
      \bottomrule
    \end{tabular}
    \subcaption{Auxiliary losses.}
    \label{tab:ablations_loss}
  \end{minipage}\vspace{-10pt}
\end{table}
 
\paragraph{Architectural components.}
Table~\ref{tab:ablations_architectural} isolates four architectural choices. Removing the bidirectional motion encoder $\mathcal{M}$ and feeding raw backbone tokens directly to the velocity DPT head (\textit{No motion branch}) drops performance by 6.2 PSNR, confirming that explicit cross-frame feature aggregation is essential for predicting consistent motion. Replacing view-dependent opacity with a single scalar (\textit{No opacity SH}) causes a comparable 7.5 PSNR drop, validating that opacity SHs serve as a learned confidence metric, masking misaligned Gaussians from viewpoints where the underlying geometry disagrees across cameras. Replacing the 2D pixel-shift + depth decomposition with direct 3D velocity prediction (\textit{No decomposition}) drops 1.5 PSNR. In this configuration, optical-flow supervision must be applied through differentiable rendering rather than directly on the predicted 2D shifts. The drop confirms that the decomposed parameterization, by allowing pixel-level flow supervision without rendering, provides a cleaner training signal than the rendered alternative. Removing both the decomposition and the optical-flow supervision simultaneously (\textit{No decomposition + no flow loss}) causes training to fail outright: predicted velocity magnitudes grow without bound and the loss diverges within the first few thousand steps, indicating that the decomposition and flow loss are jointly necessary to constrain motion predictions in our pose-free multi-view setting. Finally, removing the sinusoidal temporal tokens from the backbone (\textit{No temporal encoding}) drops 1.7 PSNR, indicating that the alternating attention layers benefit from an explicit temporal signal when aggregating features across frames.\vspace{-5pt}
 
\paragraph{Auxiliary losses.}
Table~\ref{tab:ablations_loss} ablates the three auxiliary losses on top of reconstruction. Two findings stand out. First, the consistency loss alone is catastrophic (recon + consis only, 3.51 PSNR). Without geometric grounding from distillation or flow, consistency constrains rendered depth to match backbone-predicted depth but provides no signal anchoring the geometry to the target; the model converges to a degenerate solution where Gaussians collapse together, satisfying the constraint trivially. This confirms that consistency functions as a regularizer, not a primary supervisory signal. Second, the leave-one-out comparisons reveal complementary roles: removing distillation causes the largest drop ($-3.24$ PSNR), as the alternating attention layers drift away from the pretrained DA3 representations; removing flow ($-0.96$ PSNR) costs the model dynamic awareness, with predicted velocities collapsing toward zero; and removing consistency ($-1.13$ PSNR) allows multi-view geometric disagreements to persist. The full configuration achieves the best result by combining all three, each addressing a complementary failure mode: distillation preserves priors, flow grounds dynamics, and consistency enforces multi-view coherence.
 

%% file: chapters/5_discussion.tex
\section{Conclusion}
\label{sec:conclusion}

We presented NoPo4D, a feed-forward system that fills a previously empty quadrant in 4D scene reconstruction: dynamic scenes captured by multiple synchronized cameras with unknown poses, in general environments, reconstructed in a single forward pass. The system rests on three design choices: a velocity decomposition that allows direct optical-flow supervision without rendering, a bidirectional motion encoder that aggregates features across views and time, and view-dependent opacity that mitigates the cross-view and cross-timestep Gaussian misalignments inherent to pose-free multi-view dynamic capture. Across four multi-view dynamic benchmarks NoPo4D consistently outperforms feed-forward baselines, and with brief post-optimization surpasses per-scene optimization methods that require known poses while running orders of magnitude faster.

\paragraph{Limitations and future work.}
NoPo4D assumes a static camera rig, which holds for capture setups such as Ego-Exo4D but degrades in quality when cameras themselves move. The model relies on pseudo ground-truth signals from pretrained models; as these teachers improve, NoPo4D benefits automatically without architectural changes, but training quality is bounded by their accuracy. The model also requires multiple synchronized streams; extending the same parameterization to monocular or asynchronous capture remains open. Finally, the velocity decomposition is currently first-order in time; capturing higher-order motion within the same feed-forward framework is a natural next direction. 

\paragraph{Acknowledgement}
This work was supported by the Swiss AI Initiative through a grant from the Swiss National Supercomputing Centre (CSCS) under project ID ab036 on the Alps infrastructure; the Swiss National Science Foundation (Advanced Grant 216260: "Beyond Frozen Worlds: Capturing Functional 3D Digital Twins from the Real World"); the European Union’s Horizon Europe research and innovation programme under grant agreement No. 101214398 (ELLIOT); the Horizon Europe programme under grant agreement No. 101298421 (GRAIL); and the ETH AI Center through an ETH AI Center postdoc fellowship and an ETH Zurich Career Seed Award awarded to Sunghwan Hong.

%% file: chapters/6_acknowledgement.tex

%% file: chapters/7_future-work.tex

%% file: chapters/8_bibliography.tex
\bibliographystyle{plainnat}
\bibliography{references, references2}

%% file: chapters/appendix.tex
\clearpage
\section*{Supplementary Material}
 
This document provides additional analysis supporting the main paper.
\S\ref{supp:tuning} compares different backbone fine-tuning strategies.
\S\ref{supp:alignment} details the two-pass pose alignment used for target-view rendering.
\S\ref{supp:scalability} analyzes robustness to varying input view counts.
\S\ref{supp:qualitative} presents additional qualitative comparisons across all four benchmarks.
\S\ref{supp:failure} documents failure cases.
\S\ref{supp:impact} discusses broader impacts.
 
\section{Backbone Fine-Tuning Strategy}
\label{supp:tuning}
 
The main paper notes that NoPo4D keeps the backbone's depth and camera prediction heads frozen while fine-tuning only the alternating attention layers. Here we justify this design by comparing against alternative fine-tuning configurations on ExoRecon.
 
\begin{table}[htbp]
  \caption{\textbf{Comparison of backbone fine-tuning strategies on ExoRecon.} Each row varies which components of the pretrained backbone are unfrozen during training. Fine-tuning all 16 alternating attention layers, while keeping the depth and camera heads frozen, achieves the best balance between preserving pretrained priors and adapting to the dynamic multi-view setting.}
  \label{tab:tuning}
  \centering
  \footnotesize
  \renewcommand{\arraystretch}{1.1}
  \setlength{\tabcolsep}{6pt}
  \begin{tabular}{@{}lcccc@{}}
    \toprule
    Method                          & Trainable params & PSNR $\uparrow$ & SSIM $\uparrow$ & LPIPS $\downarrow$ \\
    \midrule
    Unfreeze camera and depth heads & 525 M  & 16.94 & 0.654 & 0.354 \\
    Full fine-tuning                 & 621 M  & 25.09 & 0.829 & 0.162 \\
    Freeze backbone                 & 266 M  & 25.68 & 0.804 & 0.151 \\
    Fine-tune last 4 layers         & 318 M  & 26.32 & 0.841 & 0.152 \\
    Fine-tune last 8 layers         & 369 M  & 27.28 & 0.853 & 0.140 \\
    \midrule
    \rowcolor{oursblue}
    \textbf{Ours (fine-tune 16 alternating layers)} & 470 M & \textbf{29.15} & \textbf{0.886} & \textbf{0.125} \\
    \bottomrule
  \end{tabular}
\end{table}
 
Table~\ref{tab:tuning} reveals four findings. First, unfreezing the depth and camera heads (\textit{Unfreeze camera and depth heads}) is catastrophic, dropping PSNR by 12.4 points compared to our configuration. The pretrained DA3 heads encode strong geometric priors that are easily destabilized by the noisy gradients flowing back from the dynamic Gaussian losses; once these priors are corrupted, the unprojected point map (and therefore Gaussian means) becomes unreliable. Second, fully fine-tuning the entire backbone (\textit{Full fine-tuning}) also underperforms our configuration by 4.3 PSNR, indicating that the deeper backbone layers contain general-purpose representations that benefit from preservation. Third, completely freezing the backbone (\textit{Freeze backbone}) under-fits, dropping 3.7 PSNR; the dynamic multi-view setting requires task-specific adaptation that the frozen backbone cannot provide. Fourth, progressive unfreezing from the back of the network (\textit{Fine-tune last 4 / last 8 layers}) shows monotonic improvement as more layers are exposed, but our strategy of unfreezing all 16 alternating attention layers (while keeping the within-view layers and pretrained heads frozen) achieves the best result. This isolates the parameters that govern cross-view feature aggregation, where the dynamic multi-view setting most differs from DA3's pretraining objective.
 
\section{Pose Alignment for Target-View Rendering}
\label{supp:alignment}
 
Since NoPo4D is pose-free, rendering target views during training requires aligning predicted target poses with the reconstructed scene. We use a two-pass strategy: the first pass processes only context frames, producing Gaussians and predicted context poses that define the scene's coordinate frame. A second pass processes all frames under \texttt{no\_grad}, yielding fresh context and target poses. Umeyama alignment between the two context-pose sets gives a closed-form Sim(3) mapping target poses into the scene's frame. Because the second pass is detached, the rendering loss supervises only the Gaussians from the first pass, not the encoder's pose predictions. The same two-pass strategy is used at inference, where no ground-truth poses are available.

\section{Robustness to Varying Input View Counts}
\label{supp:scalability}

NoPo4D is trained on multi-view sequences with 1--4 cameras but can be deployed at inference time with arbitrary view counts. We evaluate this generalization by varying the number of input cameras $C \in \{2, \ldots, 12\}$ on N3DV.

\begin{figure}[htbp]
    \centering
    
    \begin{subfigure}{0.32\linewidth}
        \begin{tikzpicture}
            \begin{axis}[
                width=\linewidth, height=5.5cm,
                title={PSNR ($\uparrow$)},
                xlabel={Input Cameras}, ylabel={PSNR (dB)},
                xmin=2, xmax=12, ymin=12, ymax=32,
                xtick={2,4,6,8,10,12},
                grid=both, grid style={dashed, gray!30},
                label style={font=\footnotesize},
                tick label style={font=\footnotesize},
                title style={font=\small\bfseries},
                legend columns=-1, 
                legend entries={NoPo4D, Neoverse, Movies, DGGT},
                legend to name=commonlegend,
                legend style={draw=none, fill=none, column sep=2ex, font=\small}
            ]
            \addplot[blue, thick, mark=*] coordinates {(2,30.81)(3,29.3)(4,28.43)(5,28.38)(6,28.29)(7,27.99)(8,27.72)(9,27.44)(10,27.08)(11,26.64)(12,26.12)};
            \addplot[red, thick, mark=square*] coordinates {(2,27.9)(3,25.65)(4,24.5)(5,23.53)(6,23.08)(7,22.59)(8,22.23)(9,22.25)(10,21.76)(11,21.3)(12,20.98)};
            \addplot[orange, thick, mark=triangle*] coordinates {(2,15.11)(3,13.85)(4,14.1)(5,15.98)(6,16.07)(7,15.77)(8,15.39)(9,15.07)(10,14.86)(11,14.65)(12,14.67)};
            \addplot[violet, thick, mark=diamond*] coordinates {(2,24.44)(3,23.71)(4,23.11)(5,23.06)(6,22.95)(7,22.49)(8,22.01)(9,21.82)(10,21.76)(11,21.30)(12,20.80)};
            \end{axis}
        \end{tikzpicture}
    \end{subfigure}
    \hfill
    \begin{subfigure}{0.32\linewidth}
        \begin{tikzpicture}
            \begin{axis}[
                width=\linewidth, height=5.5cm,
                title={SSIM ($\uparrow$)},
                xlabel={Input Cameras}, ylabel={SSIM},
                xmin=2, xmax=12, ymin=0.4, ymax=1.0,
                xtick={2,4,6,8,10,12},
                grid=both, grid style={dashed, gray!30},
                label style={font=\footnotesize},
                tick label style={font=\footnotesize},
                title style={font=\small\bfseries}
            ]
            \addplot[blue, thick, mark=*] coordinates {(2,0.9255)(3,0.9148)(4,0.9057)(5,0.9028)(6,0.8987)(7,0.8911)(8,0.8856)(9,0.8776)(10,0.867)(11,0.8586)(12,0.848)};
            \addplot[red, thick, mark=square*] coordinates {(2,0.8745)(3,0.8206)(4,0.7895)(5,0.7624)(6,0.7442)(7,0.7273)(8,0.7195)(9,0.7144)(10,0.7021)(11,0.6792)(12,0.668)};
            \addplot[orange, thick, mark=triangle*] coordinates {(2,0.654)(3,0.590)(4,0.556)(5,0.524)(6,0.505)(7,0.493)(8,0.474)(9,0.465)(10,0.461)(11,0.448)(12,0.432)};
            \addplot[violet, thick, mark=diamond*] coordinates {(2,0.810)(3,0.770)(4,0.750)(5,0.750)(6,0.751)(7,0.736)(8,0.722)(9,0.713)(10,0.706)(11,0.688)(12,0.667)};
            \end{axis}
        \end{tikzpicture}
    \end{subfigure}
    \hfill
    \begin{subfigure}{0.32\linewidth}
        \begin{tikzpicture}
            \begin{axis}[
                width=\linewidth, height=5.5cm,
                title={LPIPS ($\downarrow$)},
                xlabel={Input Cameras}, ylabel={LPIPS},
                xmin=2, xmax=12, ymin=0.0, ymax=0.7,
                xtick={2,4,6,8,10,12},
                grid=both, grid style={dashed, gray!30},
                label style={font=\footnotesize},
                tick label style={font=\footnotesize},
                title style={font=\small\bfseries}
            ]
            \addplot[blue, thick, mark=*] coordinates {(2,0.1167)(3,0.1355)(4,0.1443)(5,0.1456)(6,0.1447)(7,0.1479)(8,0.1491)(9,0.1535)(10,0.1598)(11,0.1674)(12,0.1759)};
            \addplot[red, thick, mark=square*] coordinates {(2,0.2257)(3,0.284)(4,0.3122)(5,0.3334)(6,0.3448)(7,0.3503)(8,0.3536)(9,0.3612)(10,0.3703)(11,0.3835)(12,0.3908)};
            \addplot[orange, thick, mark=triangle*] coordinates {(2,0.365)(3,0.454)(4,0.505)(5,0.532)(6,0.533)(7,0.546)(8,0.561)(9,0.572)(10,0.581)(11,0.592)(12,0.602)};
            \addplot[violet, thick, mark=diamond*] coordinates {(2,0.247)(3,0.262)(4,0.270)(5,0.268)(6,0.271)(7,0.295)(8,0.303)(9,0.309)(10,0.318)(11,0.329)(12,0.345)};
            \end{axis}
        \end{tikzpicture}
    \end{subfigure}

    \vspace{0.2cm}
    \ref{commonlegend}
    \vspace{0.3cm}
    
    \begin{subfigure}{0.32\linewidth}
        \begin{tikzpicture}
            \begin{axis}[
                width=\linewidth, height=5.5cm,
                title={Peak Memory ($\downarrow$)},
                xlabel={Input Cameras}, ylabel={Memory (GB)},
                xmin=2, xmax=12, ymin=0, ymax=55, 
                xtick={2,4,6,8,10,12},
                grid=both, grid style={dashed, gray!30},
                label style={font=\footnotesize},
                tick label style={font=\footnotesize},
                title style={font=\small\bfseries}
            ]
            \addplot[blue, thick, mark=*] coordinates {(2,5.85)(3,8.24)(4,10.10)(5,13.66)(6,15.61)(7,17.59)(8,19.51)(9,21.51)(10,23.35)(11,24.89)(12,26.30)};
            
            \addplot[red, thick, mark=square*] coordinates {(2,5.89)(3,8.90)(4,10.84)(5,12.79)(6,14.72)(7,16.66)(8,18.60)(9,20.53)(10,22.47)(11,23.95)(12,25.27)};
            
            \addplot[orange, thick, mark=triangle*] coordinates {(2,9.98)(3,10.94)(4,11.90)(5,12.85)(6,13.81)(7,14.73)(8,15.70)(9,16.61)(10,17.60)(11,18.52)(12,19.48)};
            
            \addplot[violet, thick, mark=diamond*] coordinates {(2,13.66)(3,16.88)(4,20.55)(5,24.29)(6,28.05)(7,31.80)(8,35.57)(9,39.33)(10,43.08)(11,46.86)(12,50.62)};
            \end{axis}
        \end{tikzpicture}
    \end{subfigure}
    \hspace{0.05\linewidth} 
    \begin{subfigure}{0.32\linewidth}
        \begin{tikzpicture}
            \begin{axis}[
                width=\linewidth, height=5.5cm,
                title={Inference Time ($\downarrow$)},
                xlabel={Input Cameras}, ylabel={Time (s)},
                xmin=2, xmax=12, ymin=0, ymax=16, 
                xtick={2,4,6,8,10,12},
                grid=both, grid style={dashed, gray!30},
                label style={font=\footnotesize},
                tick label style={font=\footnotesize},
                title style={font=\small\bfseries}
            ]
            \addplot[blue, thick, mark=*] coordinates {(2,0.90)(3,1.01)(4,1.22)(5,1.57)(6,1.93)(7,2.42)(8,2.79)(9,3.43)(10,3.79)(11,4.48)(12,5.08)};
            
            \addplot[red, thick, mark=square*] coordinates {(2,0.81)(3,0.85)(4,1.14)(5,1.35)(6,1.65)(7,1.91)(8,2.24)(9,2.60)(10,2.82)(11,3.20)(12,3.60)};
            
            \addplot[orange, thick, mark=triangle*] coordinates {(2,0.66)(3,0.71)(4,0.89)(5,1.14)(6,1.44)(7,1.63)(8,1.91)(9,2.17)(10,2.46)(11,2.78)(12,3.12)};
            
            \addplot[violet, thick, mark=diamond*] coordinates {(2,1.17)(3,1.66)(4,2.49)(5,3.49)(6,4.67)(7,6.02)(8,7.52)(9,9.16)(10,11.03)(11,13.11)(12,15.34)};
            \end{axis}
        \end{tikzpicture}
    \end{subfigure}

    \caption{\textbf{Scalability Analysis.} Multi-view input density vs. rendering quality (top row) and computational cost (bottom row). DGGT inference time and scaling curves indicate distinct trade-offs against baseline models.}
    \label{fig:scalability}
\end{figure}
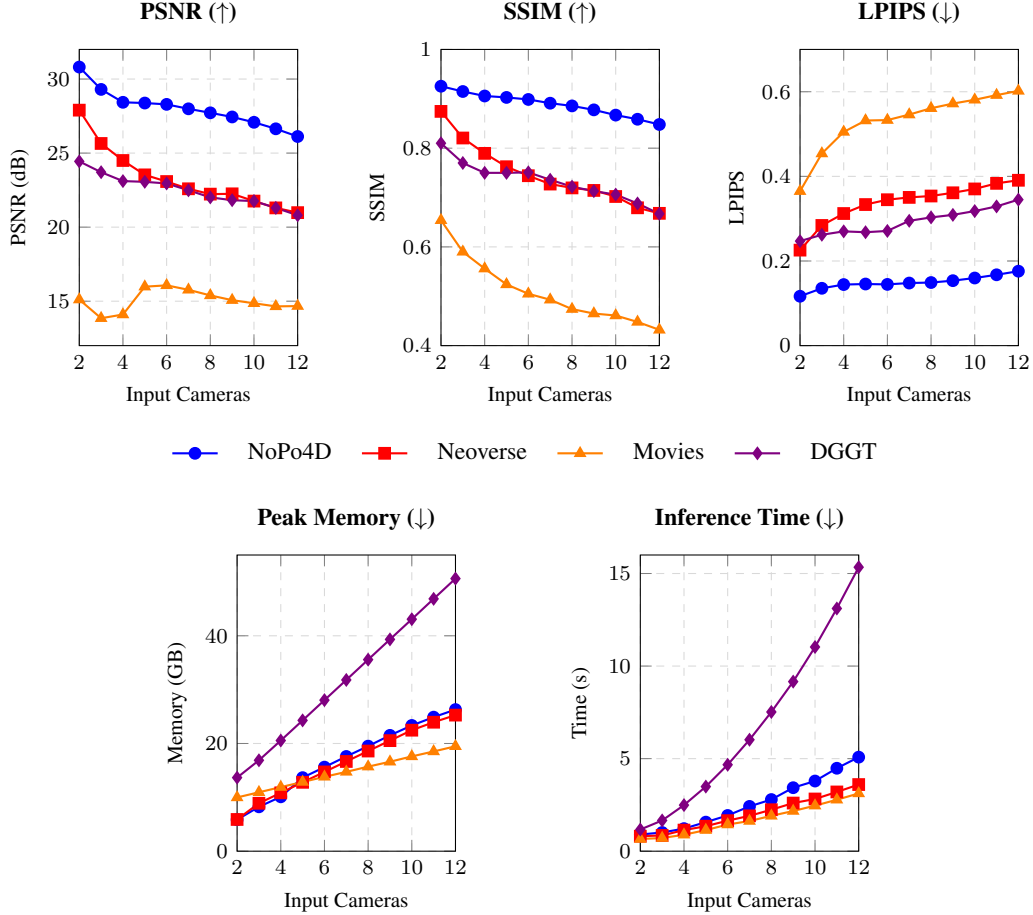

Figure~\ref{fig:scalability} reveals several findings. First, NoPo4D maintains a substantial quality lead over all baselines across the full range of input camera counts, despite never having seen $C > 4$ during training. Second, all methods show a modest quality decrease as the camera count grows: with more cameras, the target views being evaluated are further from any input view, making the task harder. NoPo4D degrades most gracefully, losing only $\sim$4.7 PSNR from $C=2$ to $C=12$, compared to $\sim$6.9 for NeoVerse and $\sim$3.6 for DGGT. Third, computational cost scales near-linearly with input cameras for NoPo4D, NeoVerse, and MoVieS, while DGGT's memory and inference time grow super-linearly, reaching $\sim$50 GB and $\sim$15 seconds at 12 cameras. NoPo4D remains under 27 GB and 5 seconds at the same scale.

\section{Additional Qualitative Results}
\label{supp:qualitative}

\paragraph{Static qualitative comparisons.}
We provide additional side-by-side comparisons against feed-forward baselines on each of the four benchmarks. 


\section{Failure Cases}
\label{supp:failure}
\begin{figure}[htbp]
    \centering
    \begin{subfigure}[b]{0.25\linewidth}
        \centering
        \includegraphics[width=\linewidth]{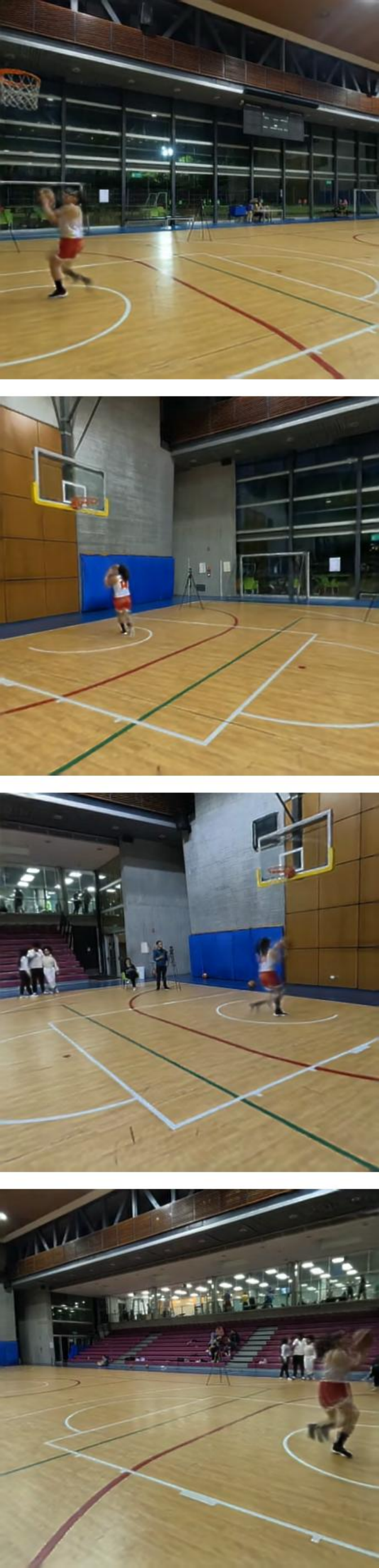}
        \caption{}
        \label{fig:failures_a}
    \end{subfigure}
    \hspace{0.07cm} 
    \begin{subfigure}[b]{0.25\linewidth}
        \centering
        \includegraphics[width=\linewidth]{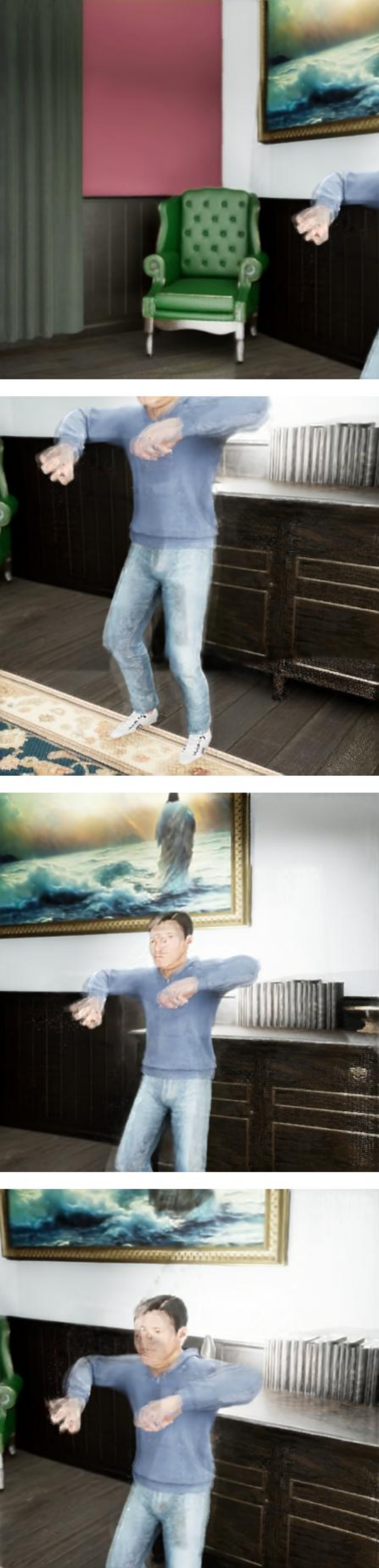}
        \caption{}
        \label{fig:failures_b}
    \end{subfigure}
    \hspace{0.07cm} 
    \begin{subfigure}[b]{0.25\linewidth}
        \centering
        \includegraphics[width=\linewidth]{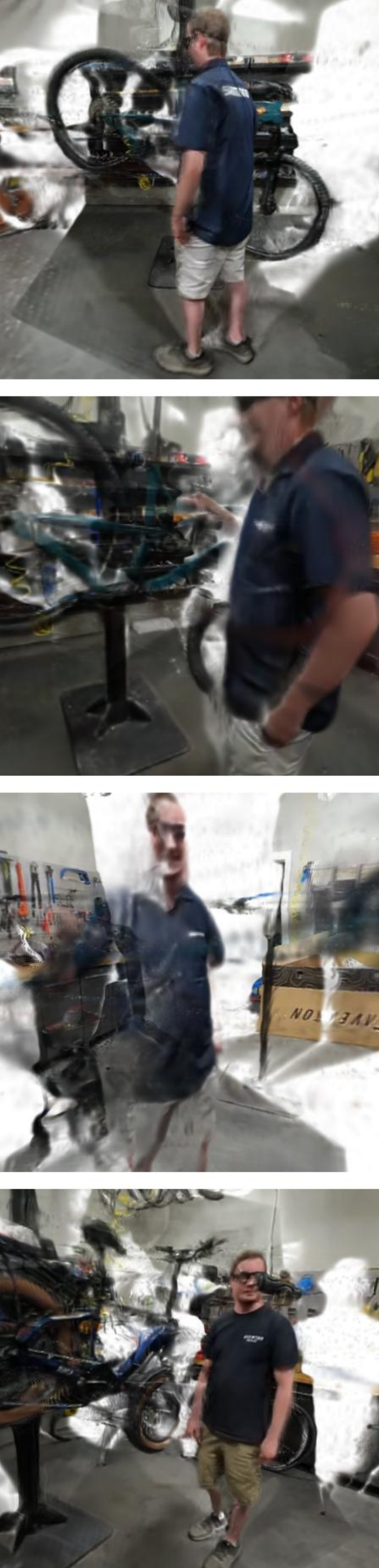}
        \caption{}
        \label{fig:failures_c}
    \end{subfigure}
    
      \caption{\textbf{Failure cases.} (a) Cross-view misalignments and floating artifacts in fast-moving regions, where the motion encoder cannot fully resolve large inter-frame displacements. (b) Degradation under camera motion (RecamMaster synthetic dataset~\cite{bai_recammaster_2025}), where the static-rig assumption causes the per-camera pose averaging to collapse distinct viewpoints into an incorrect single pose. (c) Floating Gaussian artifacts in scene regions far from input camera viewpoints, where the lack of cross-view geometric constraint leaves Gaussians at incorrect depths.}
    \label{fig:failures}
\end{figure}
We document representative failure modes in Fig.~\ref{fig:failures}.

\section{Broader Impacts}
\label{supp:impact}
 
NoPo4D advances feed-forward 4D scene reconstruction from unposed multi-view video, with potential applications~\cite{hong2021deep,hong2022cost,hong2022neural,cho2021cats,cho2022cats++,cho2024cat,shin2024towards,hong2024unifying,an_cross-view_2024,kim2025seg4diff,yoon2025visual,gropl2026entropy,lee2026tora,lee20253d,yue2025litept,han2025emergent,gurbuz2026moving} in accessible content creation, point tracking, scene understanding, correspondence estimation, robotics, sports and performance capture, egocentric scene understanding, and immersive media. We note two responsible-use considerations specific to this capability. First, reconstructing dynamic scenes from arbitrary multi-camera arrays without requiring calibration lowers the barrier to surveillance applications; downstream deployments should consider whether captured subjects have consented to 3D reconstruction, not only 2D recording. Second, the capacity to render scenes from novel viewpoints raises concerns about synthetic media when combined with generative tools. NoPo4D itself produces only reconstructions of observed content (not synthesized scenes), but practitioners building on this work should be mindful of these dual-use possibilities. We follow standard practices in our own work: training data (Ego-Exo4D) was released under its original consent and license terms, and we do not collect new human-subjects data.